\documentclass[preprint,10pt]{elsarticle}

\usepackage[title,toc,titletoc,page]{appendix}
\usepackage[font=small,labelfont=bf, justification=justified,
   format=plain]{caption}
\usepackage{subcaption}
\usepackage[a4paper,top=3cm,bottom=2cm,left=3cm,right=3cm,marginparwidth=1.75cm]{geometry}
\usepackage{amssymb}
\usepackage{amsthm}
\usepackage{algorithm}
\usepackage[noend]{algpseudocode}
\usepackage{setspace}
\usepackage{fancyhdr}
\usepackage{amsmath}
\usepackage{graphicx}
\usepackage[colorinlistoftodos]{todonotes}
\usepackage[colorlinks=true, allcolors=blue]{hyperref}
\usepackage[T1]{fontenc}
\usepackage{dsfont}
\date{}

\usepackage{etoolbox}
\makeatletter
\patchcmd{\ps@pprintTitle}
  {Elsevier}
  {arXiv}
  {}{}
\newtheorem{prop}{Proposition}
\makeatother

\begin{document}
\begin{frontmatter}
\title{Scalar Quantization as Sparse Least Square Optimization}

\author[add1,add2]{Chen Wang\fnref{t1}}
\author[add1]{Xiaomei Yang}
\author[add3]{Shaomin Fei}
\author[add1]{Kai Zhou}
\author[add1]{Xiaofeng Gong}
\author[add1]{Miao Du}
\author[add1]{Ruisen Luo\corref{cor1}}
\address[add1]{College of Electrical Engineering, Sichuan University, Chengdu, Sichuan 610065, China}
\address[add2]{Department of Computer Science, Rutgers University -- New Brunswick, Piscataway, New Jersey 08854, USA}
\address[add3]{Engineering Practice Center, Chengdu University of Information Technology, Chengdu, Sichuan 610059, China}
\cortext[cor1]{the Corresponding Author, email: rsluo@scu.edu.cn}
\fntext[]{This work is supported by Natural Science Foundation of China (NSFC 51475391), Research Project of State Key Laboratory of Southwest Jiaotong University (No.TPL1502) and University-Enterprise Cooperation Project (17H1199).}
\fntext[t1]{Work was done at Sichuan University. The author is now based on Rutgers University.}

\begin{abstract}
\normalsize
Quantization can be used to form new vectors/matrices with shared values close to the original. In recent years, the popularity of scalar quantization for value-sharing application has been soaring as it has been found huge utilities in reducing the complexity of neural networks. Existing clustering-based quantization techniques, while being well-developed, have multiple drawbacks including the dependency of the random seed, empty or out-of-the-range clusters, and high time complexity for large number of clusters. To overcome these problems, in this paper, the problem of scalar quantization is examined from a new perspective, namely sparse least square optimization. Specifically, inspired by the property of sparse least square regression, several quantization algorithms based on $l_1$ least square are proposed. In addition, similar schemes with $l_1 + l_2$ and $l_0$ regularization are proposed. Furthermore, to compute quantization results with given amount of values/clusters, this paper designed an iterative method and a clustering-based method, and both of them are built on sparse least square. The paper shows that the latter method is mathematically equivalent to an improved version of k-means clustering-based quantization algorithm, although the two algorithms originated from different intuitions. The algorithms proposed were tested with three types of data and their computational performances, including information loss, time consumption, and the distribution of the values of the sparse vectors, were compared and analyzed. The paper offers a new perspective to probe the area of quantization, and the algorithms proposed can outperform existing methods especially under some bit-width reduction scenarios, when the required post-quantization resolution (number of values) is not significantly lower than the original number.
\end{abstract}
\end{frontmatter}
\section{Introduction}
Quantization can reduce the number of bits to represent vectors and matrices by producing their replacements with shared values and acceptable differences. The technique has been found great usefulness in some areas, e.g., image processing\cite{QuantizationImageProcessing}, speech recognition\cite{QuantizationSpeechRecognition}, and machine Learning techniques\cite{QuantizationNearestNeighbour}. And recently, with the growing research interests in deploying neural networks on resource-scarce edge devices, quantization techniques have grasped considerable attention because of its ability in reducing the size of network\cite{SoftWeightSharing2011}\cite{SongHan2015NetworkCompression}\cite{EfficientProcessingDL_Tutorial}\cite{gong2014compressing}. Recent works like \cite{gong2014compressing} and \cite{SongHan2015NetworkCompression} suggest that by simply running scalar quantization, the size of the neural network can be considerably reduced, while the reduction in precision is almost negligible. A recent study \cite{EmbeddedDeepLearningThesis} (by the authors) discusses to apply novel methods quantization in neural network compression, and it can serve as the inspiration of this paper. However, the method proposed in this paper can be used in both neural network compression and general-purpose quantization.\par
Popular scalar quantization methods often adopt a clustering-based scheme\cite{EfficientProcessingDL_Tutorial}, and K-means clustering quantization and its variations are the most prominent techniques used in the area. While straightforward and convenient, these methods frequently suffer from the several problems: 1. empty clusters or irrational values (say, out-of-range values) due to bad random initialization; 2. reliance on randomness. Practical K-means clustering is usually solved by Lloyd's-like algorithms, which are heuristic methods and cannot guarantee the optimal solution; and 3. time consumption. To produce reliable results, K-means algorithm is usually executed multiple times (usually 5 to 10 times) with different initializations. While such technique could improve the quality of the results, the time complexity of this method is high, especially when the number of post-quantization values is large (high-resolution).\par
In this paper, the quantization algorithms are examined from another perspective: sparse least square optimization. The idea is straightforward to understand: if one regards each value in the original vector to be a combination of some 'basis', then by introducing sparsity to the possible amount of combinations, the number of values to be 'generated' will be constrained. Specifically, we consider the sparse-inducing properties of $l_0$ and $l_1$ norm-regularization, and design algorithms that could minimize the reconstruction difference and introduce sparsity to the 'basis' simultaneously. Based on the above idea, the $l_1$-constrained least square form of quantization algorithm is first proposed in the paper. In addition, to optimize the performance based on its computational properties, an alternative version with both $l_1$ and $l_2$ norm-regularization is explored and implemented. Furthermore, to design a least square quantization method that could produce results by indicating certain amounts of quantization amounts (instead of the value of penalization coefficient $\lambda$), two additional algorithms are designed: the first one is introduced by following the idea of the $l_1$ least square optimizations with iteratively enhancing constraints, and the second one is accomplished by combining k-means clustering with least square optimization. Interestingly, the second approach could also be interpreted as an improvement of the conventional k-means clustering quantization method. The proposed methods are compared with quantization techniques based on k-means clustering, Mixture of Gaussian, and a novel data transformation-based clustering method proposed in \cite{azimi2017novelClustering}. Experimental results illustrate that the performance of the proposed algorithms are competitive. For the $l_1$-based algorithms, they are especially favorable in terms of running-time complexity, which makes them particularly useful when the required quantization amounts is not in a trivial scale. Moreover, the results provided by the proposed $l_1$ methods are more exact and relatively more independent from random seed. \par 
Notice that our methods are very similar to sparse compression in signal processing\cite{SparseCompressionTypical}. Nevertheless, they deals with different problems: in quantization problem, the constructed vector should have shared values; while in sparse signal processing, it only demands the sparse vector to be able to produce a vector close to the original signal. \par
The rest of the paper is arranged as follows: section 2 will be introducing related work in the field; section 3 will be introducing our designed methods mathematically, and analyze their optimization properties with proofs on convergence and complexity; the experimental results of the methods are shown, compared and analyzed in section 4; and finally, a conclusion is drawn in section 5 and future research topics related to this paper is discussed. 
\section{Related Work}
Scalar quantization is a classical problem, and methods to carry out this task are relatively well-developed. \cite{EfficientProcessingDL_Tutorial} provides a brief survey for basic methods concerning quantization, which includes domain-based hand-coding methods and clustering-based techniques such as k-means quantization. The idea of quantizing vectors with clustering methods provides us an open skeleton, in which we can plug novel clustering techniques to produce new quantization algorithms \cite{EmbeddedDeepLearningThesis}. For instance, a classic study \cite{xiang1994color} adopts agglomerative clustering to perform image quantization, and achieved competitive performance at that time. Similarly, authors of \cite{hsieh2000adaptive} designed an adaptive clustering method specifically fitting their quantization technique. It is noticeable that the term 'k-means quantization' is sometimes referred as 'hard c-means quantization' in related research \cite{wen2011hard}, and a similar concept 'fuzzy c-means quantization' refers to a variation of the method that assign a 'fuzzy partition/membership parameter' to each data point \cite{le2011fast}. A key difference between fuzzy c-means and k-means quantization methods is that during the clustering procedure, each data point will contribute to the update of every cluster in the former, while will only affect one specific cluster in the latter. Moreover, after getting the clustering results, the membership of a data point will be obvious in k-means, while in fuzzy c-means the membership should be computed by taking \textit{argmax} operation. \par 
Apart from clustering-based quantization, \cite{1992weightsharing} offers an alternative technique to use Mixture of Gaussian method to perform quantization specifically for neural networks, and a recent paper \cite{2017_compression_1992idea} re-examined this idea and formally designed it to be used for neural network compression and provided a mathematical justification for it. Other techniques to perform vector quantization include \cite{DivergenceQuantization}, which utilized divergence instead of distance metric as the measurement and derived an algorithms based on it; \cite{UseNeuralNetworkToQuantize}, which designed a neural network to perform quantization; and \cite{PairwiseSimilarityQuantization}, which considered pairwise dis-similarity as the optimization metric. Moreover, quantization is of practical usage in areas of signal processing and image retrieval with specific constraints and characteristics. Thus, there are scalar quantization algorithms specifically designed to optimize evaluation metrics of the fields \cite{boufounos2012QuantizationForSignalProcessing,zhou2016QuantizationForImageRetrieval}. Furthermore, for high-dimensional vector quantization, there have been multiple codebook-based algorithms with various optimization strategies \cite{ai2017ResidualVectorQuant,ozan2016CompetitiveVectorQuant}. However, despite these developments in the area, to the best of the authors' knowledge, hitherto there has not been publication discussing scalar quantization algorithms as sparse least square optimization. \par
There are notable amount of academic publications discussing the applications of vector quantization, and recently, research lying in this area has been connected to neural networks, as the ability of quantization in compressing model size is being exploited in implementing neural networks on edge devices. \cite{gong2014compressing} conducted a notable study of implementing quantization to the compression of neural network, and illustrated that simply applying k-mean scalar quantization can achieve satisfying compression rate while maintaining an acceptable accuracy. Similarly, in a more recent study, \cite{guo2018survey} provides a survey for the usage of quantization in neural networks, and listed the major challenges in the area. \cite{SongHan2015NetworkCompression} proposed a general pipeline to reduce model storage, and an important part of it is quantization. And similar with \cite{2017_compression_1992idea} mentioned above, \cite{FixedPointQuantizationNetwork} specifically designed a fixed-point quantization method to compress neural networks. More recent work also focused on introducing weight quantization into the overall neural network field, instead of only quantizing pre-trained networks: \cite{hubara2017quantizationDuringTraining} proposed a novel network that could perform weight quantization during the training process; \cite{agustsson2017quantizationrepresentation} designed an algorithm to learn compressible representations with neural networks based on quantization; and \cite{carreira2017model} proposed a method to use codebook-based quantization to compress the neural network and learn the codebook and network parameters simultaneously. As mentioned in the introductory section, the inspiration of this work comes from a neural network weight-sharing problem mentioned in \cite{EmbeddedDeepLearningThesis}, and the quantization of a set of neural network weight parameters is tested in the experimental section. \par
The algorithm proposed in this work has significant similarity with compressive sensing (sparse signal processing) in terms of regularization idea and optimization target functions \cite{SparseCompressionTypical,SparseCompressionTypical2,SparseCompressionTypical3}. Typical approaches to induce sparsity in compressive sensing algorithms are to introduce $l_0$ norm\cite{l0SignalProcessing}, $l_1$ norm \cite{LassoRetirevePaper,LassoCompressiveSensing} and/or $l_{2,1}$ norm \cite{l21CompressiveSensing} to the target optimization functions. And similarly, our algorithms utilize these techniques to induce sparsity. Meanwhile, since $l_1$ norm is not everywhere differentiable and $l_0$ norm is not even convex, there also exist plenty of algorithms devoted to efficiently solve the optimization problems \cite{l1OptimizationAlgorithm,l0OptimizationAlgorithm,CoordinateDescentPaper}. In this paper, we use coordinate descent method for $l_1$ optimization, and the newly-proposed Fast Best Subset Selection \cite{l0Learn2018Hazimeh} (will be called '$l_0$ learn' in this paper) to optimize $l_0$ target functions.
\section{Quantization Algorithms}
\subsection{Problem Setup}
The quantization task can be described as follows: suppose we have a vector $\boldsymbol{w}$ that has $m$ distinct values. Now we intend to find a vector $\boldsymbol{w}^{*}$ with $p$ distinct values, where $p \leq m$. In some case, it can also be set as a more strict condition $p \leq l$, where $l$ is the upper bound of the post-quantization amounts of values and $l<m$. Then by denoting the differences between the original vector and the constructed one with $l_2$ norm, our original target function could be formed as:\\
\begin{equation}
\label{originalOptTargetFunc}
\begin{aligned}
& \underset{\boldsymbol{w}^{*}}{\text{minimize}}
& & ||\boldsymbol{w}-\boldsymbol{w}^{*}||_{2}^{2} \\
& \text{subject to}
& & \boldsymbol{o}(\boldsymbol{w}^{*}) \leq l \\
\end{aligned}
\end{equation}
Where $\boldsymbol{o(\cdot)}$ means the number of distinct values of the vector. And notice that here we only consider $\boldsymbol{w}$ in 1-dimension vector form (scalar quantization). If the data is coded in a matrix, such as neural network parameters and images, we can simply 'flatten' the matrix into a vector to perform quantization, and then turn it back to the original shape. However, in this paper, the methods cannot perform quantization operating directly on vectors. The design of such methods will be an interesting research topic in the future.
\subsection{Quantization as Sparse Least Square and Algorithms with $l_1$ Regularization}
To begin with, we first pre-process $\boldsymbol{w}$ into $\boldsymbol{\hat{w}}=\texttt{unique}(\boldsymbol{w})$, which we directly operate on distinct values and recover the full vector by indexing later. And to fulfill the purpose of quantization, we will be needing to construct a new vector with length $m$ and $p$ distinct values. We could assume there exist a 'base' vector $\boldsymbol{v}$ with shape $[k \times 1]$, where $k$ is a given number, and the $\boldsymbol{\hat{w}}$ is generated by $\boldsymbol{v}$ through linear transformation. Notice that there should be $k \geq m$, as it will otherwise be unreasonable to project vector in $R^{k}$ to $R^{m}$ with linear transformation. Then suppose the linear transformation matrix is $\boldsymbol{\Psi}$ (with shape $[m \times k]$), the relationship between $\boldsymbol{\hat{w}}$, $\boldsymbol{\Psi}$, and $\boldsymbol{v}$ will be:\\
$$
\boldsymbol{\hat{w}} = \boldsymbol{\Psi}\boldsymbol{v}
$$
And combining this expression with equation \ref{originalOptTargetFunc}, we can get the new optimization target:\\
\begin{equation}
\label{PsiOptTargetFunc}
\underset{\boldsymbol{\Psi},\boldsymbol{v}}{\text{min}}||\boldsymbol{\hat{w}}-\boldsymbol{\Psi}\boldsymbol{v}||_{2}^{2}
\end{equation}
Solving \ref{PsiOptTargetFunc} is a matrix decomposition problem without any form of sparsity/value-sharing. To introduce sparsity, here we introduce another matrix $\boldsymbol{A}$, with $\boldsymbol{\Psi}^{*} = \boldsymbol{A}\boldsymbol{\Psi}$, and each entry of $\boldsymbol{\Psi}^{*} $ should be:\\
\begin{equation}\label{PsiEntry}
\boldsymbol{\Psi}^{*}_{i,j} = \sum_{c=0}^{i}\alpha_c \boldsymbol{\Psi}_{c,j}
\end{equation}
By designing the matrix with this addition form, we could be able to achieve 'same values' when there exist $\alpha_c = 0$. Equation \ref{PsiEntry} could be achieved by designing matrix $\boldsymbol{A}$ as a lower-triangular matrix (with main diagonal on):\\
$$
\boldsymbol{A} = 
\begin{bmatrix}
\alpha_1 & 0 & 0 & \cdots & 0\\
\alpha_1 & \alpha_2 & 0 & \cdots & 0\\
\alpha_1 & \alpha_2 & \alpha_3 & \cdots & 0\\
\cdots & \cdots & \cdots & \cdots & \cdots \\
\alpha_1 & \alpha_2 & \alpha_3 & \cdots &  \alpha_n
\end{bmatrix}
$$
And now we have two matrices, $\boldsymbol{A}$ and $\boldsymbol{\Psi}$, to control the constructed vector. Intuitively, if we add $l_1$ and/or $l_0$ norm regularization on the target function, it will be possible for us to produce vector with shared values. Here we consider $l_1$ in the first place because it is continuous and convex. Our optimization target then become:\\
\begin{equation}\label{AlphaPsiOptimisation}
\begin{aligned}
&\min_{\boldsymbol{\alpha},\boldsymbol{\Psi}} ||\boldsymbol{\hat{w}}-\boldsymbol{w}^{*}||_{2}^{2} + \lambda ||\boldsymbol{\alpha}||_{1}\\
=&\min_{\boldsymbol{\alpha},\boldsymbol{\Psi}} ||\boldsymbol{\hat{w}}-\boldsymbol{A}\boldsymbol{\Psi}\boldsymbol{v}||_{2}^{2} + \lambda ||\boldsymbol{\alpha}||_{1}
\end{aligned}
\end{equation}
Where $\boldsymbol{\Psi}$ and $\boldsymbol{v}$ are the transformation matrix and base vector from equation \ref{PsiOptTargetFunc}, and $\boldsymbol{\alpha}$ refers to a vector with $[\alpha_{1},...,\alpha_{n}]$ values. Now the property of sparsity would be able to introduced if \ref{AlphaPsiOptimisation} is optimized. However, the there are two target matrices in the target function, which makes the optimization problem difficult. To determine the system in a convenient way, we will be needing some approximations. Here, we will fix $\boldsymbol{\Psi}$ and only optimize $\boldsymbol{\alpha}$. Now suppose $k=m$, we could pose the matrix as follows:\\
$$
\boldsymbol{\Psi} = 
\begin{bmatrix}
1 & 0 & 0 & \cdots & 0 & 0\\
-1 & 1 & 0 & \cdots & 0 & 0\\
0  & -1 & 1 & \cdots & 0 & 0\\
\cdots & \cdots & \cdots & \cdots & \cdots & \cdots \\
0 & 0 & 0 & \cdots & -1 & 1
\end{bmatrix}
$$
And we will get a $[m \times 1]$ vector $\boldsymbol{v}^{*}$, in the following format:\\
$$
\boldsymbol{v}^{*} = 
\begin{bmatrix}
v_{1}\\
v_{2}-v_{1}\\
v_{3}-v_{2}\\
\cdots\\
v_{k} - v_{k-1}
\end{bmatrix}
$$
Since the the transformation matrix $\boldsymbol{\Psi}$ is within full rank, the linear space $R^{k}$ is not changed. This property implies that with proper configuration, we can find the correct solution of $\boldsymbol{\alpha}$. Notice that the above transformation matrix is given under the assumption of $k=m$, and for the $k>m$ scenario, we can simply leave some of the rows of $\boldsymbol{\Psi}$ as $0$ and keep the rank as $m$.\\
After the above transformations, the optimization target now becomes:\\
\begin{equation}
\min_{\boldsymbol{\alpha}} ||\boldsymbol{\hat{w}}-\boldsymbol{A}\boldsymbol{v}^{*}||_{2}^{2} + \lambda ||\boldsymbol{\alpha}||_{1}
\end{equation}
And this is equivalent to form a vector of $\boldsymbol{\alpha}$ and a lower-triangular matrix $\boldsymbol{V}$:\\
$$
\boldsymbol{V} = 
\begin{bmatrix}
v_{1} & 0 & 0 & \cdots & 0\\
v_{1} & v_{2}-v_{1} & 0 & \cdots & 0 \\
v_{1} & v_{2}-v_{1} & v_{3}-v_{2} & \cdots & 0\\
\cdots & \cdots & \cdots & \cdots & \cdots \\
v_{1} & v_{2}-v_{1} & v_{3}-v_{2} & \cdots & v_{m}-v_{m-1}
\end{bmatrix}
$$
And in practice, we simply use the value of original unique-value vector $\boldsymbol{\hat{w}}$ to fill the value of $\boldsymbol{v}^{*}$. The configuration of matrix $\boldsymbol{V}$ ensures global convergence and convenience in finding initial values, as one can see a further discussion in section \ref{subsubsec:convergence}. The final optimization target with $l_1$ regularization will be as follows:\\
\begin{equation}\label{l1normReguliriseOpt}
\min_{\boldsymbol{\alpha}} ||\boldsymbol{\hat{w}} -\boldsymbol{V}\boldsymbol{\alpha}||_{2}^{2} + \lambda ||\boldsymbol{\alpha}||_{1}
\end{equation}
Equation \ref{l1normReguliriseOpt} is very similar to the optimization target in compressive sensing. Nevertheless, there are two significant differences: firstly, the root of the target function and the derivations are different from those in compressive sensing; and secondly, the produced vector $\boldsymbol{w}^{*}$ will be a quantized vector instead of just a sparse vector close to the original as in compressive sensing. \par
By introducing sparsity, target function \ref{l1normReguliriseOpt} loses its property of being everywhere differentiable. Thus, the solution cannot usually be found analytically, and one has to use numerical (often gradient/proximal-based) methods to find the solution. This will inevitably increase the time complexity, and the exact time cost will be discussed in the later passages. However, although an analytical solution becomes unable to obtain, the optimization of the target function is not uncommon: it is a typical LASSO problem. In this paper, we employ \textbf{Coordinate Descent} method to solve it, and an argument of linear and global convergence has been put in section \ref{subsubsec:convergence}. In the experiments, the LASSO solvers we used is based on the program in Sk-learn \cite{scikit-learn}. \par
It is noticeable that the 'raw result' of equation \ref{l1normReguliriseOpt} can still be improved. As the optimization should satisfy both sparsity and $l_2$ loss, the values in the solved $\boldsymbol{\alpha}$ vector might not optimally reduce the difference between $\boldsymbol{\hat{w}}$ and the constructed vector. Thus, we consider to \textbf{solve the least square with positions leading to $\alpha \neq 0$} to further improve the result. Mathematically, this idea can be denoted as to use matrix $\boldsymbol{V}^{*}$ to perform the least square optimization, where the $\boldsymbol{V}^{*}$ matrix should be:\\
\begin{equation}
\label{NpStarRetrievalMethod}
\boldsymbol{V}^{*}_{\cdot,j} = \boldsymbol{V}_{\cdot,h_{j}},\text{ } \forall \text{ } h_{j} \text{ such that } \alpha_{h_{j}} \neq 0
\end{equation}
Which means, the $\boldsymbol{V}^{*}$ will pick the columns with corresponding non-zeros indexes in $\boldsymbol{\alpha}$. The optimization target will therefore be:\\
\begin{equation}
\label{LeastSqaurePostLassoOptimization}
\min_{\boldsymbol{\hat{\alpha}}} ||\boldsymbol{\hat{w}} -\boldsymbol{V}^{*}\boldsymbol{\hat{\alpha}}||_{2}^{2}
\end{equation}
The target function \ref{LeastSqaurePostLassoOptimization} is in a everywhere-differentiable least-square form, thus it could be direct solved analytically:\\
\begin{equation}
\label{PostLassoLeastSquareSolution}
\boldsymbol{\hat{\alpha}}^{*} = ({\boldsymbol{V}^{*}}^{T}{\boldsymbol{V}^{*}})^{-1}{\boldsymbol{V}^{*}}^{T}\boldsymbol{\hat{w}}
\end{equation}
Where $\boldsymbol{\hat{\alpha}}^{*}$ would be $[h \times 1]$ vector, where $h$ is the number of distinct values. The values of could be put back into the $\boldsymbol{\alpha}$ vector to get the final result $\boldsymbol{\alpha}^{*}$:\\
\begin{equation}
\label{FinalAlphaComputation}
\boldsymbol{\alpha}^{*}_{i} = 
\begin{cases}
\boldsymbol{\hat{\alpha}}^{*}_{h_{i}}, \boldsymbol{\alpha}_{i} \neq 0 \\
0, \text{ else}
\end{cases}
\end{equation}
And finally, the quantized vector could be constructed by multiplying the $\boldsymbol{\alpha}^{*}$ vector with the 'based transformation' matrix $\boldsymbol{V}$:\\
\begin{equation}
\label{QuantizedVectorComputing}
\boldsymbol{w}^{*} = \boldsymbol{V}\boldsymbol{\alpha}^{*}
\end{equation}
The overall quantization method with $l_1$ regularization could be denoted as algorithm \ref{alg:l1Quantization}. In the experiment section, the results of $l_1$-based algorithms with and without least square to optimize $\boldsymbol{\hat{\alpha}}^{*}$ will be shown separately.
\begin{algorithm}
\caption{Quantization with $l_1$ Least Square}\label{alg:l1Quantization}
\hspace*{\algorithmicindent} \textbf{Input: }\text{Original vector $\boldsymbol{w}$} \\
\hspace*{\algorithmicindent} \textbf{Output: }\text{Quantized vector $\boldsymbol{w}^{*}$} 
\begin{algorithmic}[1]
\State $\boldsymbol{\hat{w}} \gets \texttt{unique}(\boldsymbol{w})$
\State Optimize target function \ref{l1normReguliriseOpt} with Coordinate Descent, get $\boldsymbol{\alpha}$
\State Retrieve $\boldsymbol{V}^{*}$ with equation \ref{NpStarRetrievalMethod}
\State Compute $\boldsymbol{\hat{\alpha}}^{*}$ with equation \ref{PostLassoLeastSquareSolution}
\State Compute $\boldsymbol{\alpha}^{*}$ vector with equation \ref{FinalAlphaComputation}
\State Compute the desired $\boldsymbol{w}^{*}$ vector with equation \ref{QuantizedVectorComputing}
\end{algorithmic}
\end{algorithm}

\subsubsection{Convergence of the Optimization Target}
\label{subsubsec:convergence}
Algorithm \ref{alg:l1Quantization} achieves sparsity and quantization through the $l_1$-regularized least square form in equation \ref{l1normReguliriseOpt}. In general, the convergence of such type of target function is not guaranteed. However, as we will discuss in this section, with the Coordinate Descent method applied in this paper, the optimization target in equation \ref{l1normReguliriseOpt}  will linearly converge to a global minimum. In addition, by showing the convergence of algorithm \ref{alg:l1Quantization}, the convergence of other $l_1$-based algorithm mentioned in this paper can be analyzed in a similar manner. The convergence of target function \ref{l1normReguliriseOpt} is based on the following result:
\begin{prop}
The target function \ref{l1normReguliriseOpt} is strongly (and strictly) convex.
\end{prop}
\begin{proof}
The proposition can be verified by showing both the least square and the regularization parts are strictly convex. Showing the $l_1$-regularization part is strongly convex is trivial; For the least square part, consider computing the Hessian with respect to $\boldsymbol{\alpha}$, which will result in $\boldsymbol{H^{\alpha}} = \boldsymbol{V}^{T}\boldsymbol{V}$. Let $\boldsymbol{V}_{(i)} := \boldsymbol{V}_{i,i}$ and $v_{0} = 0$, the Hessian matrix will be: 
\begin{equation}
\label{equ:HessianOfLeastSquare}
\begin{split}
\boldsymbol{H^{\alpha}} & = \min\{n-i+1, n-j+1\}(\boldsymbol{V}_{(i)}\boldsymbol{V}_{(j)})\\
& = \min\{n-i+1, n-j+1\}(v_{i}-v_{i-1})(v_{j}-v_{j-1})
\end{split}
\end{equation}
Now notice that $\boldsymbol{V}$ is of full column rank (since $\boldsymbol{V}_{(i)} \neq 0, \text{ } \forall i$), by definition there will be $\boldsymbol{z}^{T}\boldsymbol{H^{\alpha}}\boldsymbol{z}=\boldsymbol{z}^{T}\boldsymbol{V}^{T}\boldsymbol{V}\boldsymbol{z}>0$ for all non-zero vectors $z \in R^{m}$. Thus, the Hessian matrix is positive definite, and the least square part is strongly convex. And finally, the summation of two strongly convex functions will result in a strongly convex function.
\end{proof}
The proposition has two important implications regarding the optimization of the target function \ref{l1normReguliriseOpt}: 1. Since the function is strongly convex, there \textbf{exists one (and only one) global optimum}; and 2. Coordinate Descent algorithm \textbf{converges linearly} to it (\cite{luo1993error,hong2017ConvergenceCoordinate}). These two properties bring up favorable optimization characteristics for algorithm \ref{alg:l1Quantization}, especially when one compares to the unstable optimization dynamic of k-means. \par
Moreover, from a qualitative analysis perspective, the configuration of matrix $\boldsymbol{V}$ further provides a straightforward way to set initial value for optimization. By simply setting $\boldsymbol{\alpha}_{0}=\mathds{1}^{m}$, the least square part will be of $0$ loss. Starting from this point, it is straightforward to show that setting a single value $\alpha_{i}=0$ and optimizing $\alpha_{i-1}$ can limit the square loss to $(v_{i}v_{i-1}-\frac{(v_{i}^{2}+v_{i-1}^{2})}{2})$. Furthermore, when using Coordinate Descent to optimize one dimension of the $\boldsymbol{V}$ matrix, other dimensions will get optimized towards the converging direction at the same time. Consequently, the convergence characteristics of the target function \ref{l1normReguliriseOpt} under the setup of this paper should be preferable.

\subsection{$l_1+l_2$ Regularization Algorithm and $l_0$ Regularization Algorithm}
One possible improvement of algorithm \ref{alg:l1Quantization} will be to add a \textbf{negative $l_2$ penalization term} to the original. The optimization target can be denoted with the following formula:
\begin{equation}\label{l1l2normReguliriseOpt}
\min_{\boldsymbol{\alpha}} ||\boldsymbol{\hat{w}} -\boldsymbol{V}\boldsymbol{\alpha}||_{2}^{2} + \lambda_1 ||\boldsymbol{\alpha}||_{1} - \lambda_2 ||\boldsymbol{\alpha}||_{2}^{2}
\end{equation}
Equation \ref{l1l2normReguliriseOpt} is similar with Elastic Net \cite{zou2005ElasticNet}, but with a negative $l_2$ coefficient. The intuition behind this scheme is that $l_1$ optimization often leads to $\alpha$ values with small quantities before it could reach $0$. Thus, adding the 'negative $l_2$ norm' can be regarded as a relaxation for the original $l_1$ least square to find sparse index while keep the non-zero values on their original level. More formally, if we inspect the mathematical expressions under coordinate descent with shrinkage model, the LASSO optimization could be expressed as:\\
\begin{equation}
\label{LassoCoordinateDescent}
\alpha_{k}^{t+1} = S_{\frac{\lambda_{1}}{\boldsymbol{V}_{\cdot,k}^{T}\boldsymbol{V}_{\cdot,k}}}(\frac{\boldsymbol{V}_{\cdot,k}^{T}\boldsymbol{\hat{w}}-\boldsymbol{V}_{\cdot,k}^{T}\boldsymbol{V}_{\cdot,/k}\boldsymbol{\alpha}_{/k}^{t}}{\boldsymbol{V}_{\cdot,k}^{T}\boldsymbol{V}_{\cdot,k}})
\end{equation}
Where $k$ denotes the coordinate to be optimized and $S_{a}(x)$ means the shrinkage operator defined by:\\
\begin{equation*}
S_{\lambda}(x) = 
\begin{cases}
x-\lambda, x\geq \lambda\\
x+\lambda, x\leq -\lambda\\
0, -\lambda < x < \lambda
\end{cases}
\end{equation*}
for positive-valued $\lambda$. In comparison, the Coordinate Descent for negative $l_2$ penalization will be:\\
\begin{equation}
\label{l1l2CoordinateDescent}
\alpha_{k}^{t+1} = S_{\frac{\lambda_{1}}{\boldsymbol{V}_{\cdot,k}^{T}\boldsymbol{V}_{\cdot,k}-2\lambda_2}}(\frac{\boldsymbol{V}_{\cdot,k}^{T}\boldsymbol{\hat{w}}-\boldsymbol{V}_{\cdot,k}^{T}\boldsymbol{V}_{\cdot,/k}\boldsymbol{\alpha}_{/k}^{t}}{\boldsymbol{V}_{\cdot,k}^{T}\boldsymbol{V}_{\cdot,k}-2\lambda_2})
\end{equation}
Which means, for the $l_1 + l_2$ combined optimization, the proximal been projected will be larger as the denominator of will be subtracting a positive value. Also, the absolute value of the threshold to be shrinkaged as $0$ is higher, making the vector easier to achieve sparsity. There are rare, if any, integrated Lasso optimization packages that permits the parameter setting like equation \ref{l1l2normReguliriseOpt}. Thus, this algorithm is optimized by the coordinate descent method implemented by the authors. \par
Another variation of the algorithm based on \ref{l1normReguliriseOpt} could be to replace the $l_1$ norm with $l_0$ norm. In the $l_0$ algorithm, instead of directly add a penalization term, we explicitly set limitations of the number of distinct values:
\begin{equation}\label{l0normReguliriseOpt}
\begin{aligned}
& \underset{\boldsymbol{\alpha}}{\text{min}}
& & ||\boldsymbol{\hat{w}} -\delta\boldsymbol{V^{*}}\boldsymbol{\alpha}||_{2}^{2} \\
& \text{subject to}
& & ||\boldsymbol{\alpha}||_{0} \leq l \\
\end{aligned}
\end{equation}
Where $l$ is a number that we manually set, which indicate the upper bound of the amount of distinct values. Finding the exact solution with $l_0$ norm is NP-hard \cite{L0NPhardPaper}, thus we could only be solve by heuristic-based algorithms up to now. In this paper we utilize the recent-proposed 'L0Learn' as mentioned above \cite{l0Learn2018Hazimeh}, which utilizes combinational optimization scheme with coordinate descent and support the value of $l$(amount of quantization) up to 100. However, one should notice that the $l_0$ optimization method is not universal, which means, it could not reach arbitrary required number of values under our settings. Also, this is the reason why one can only specify an 'upper bound' ($l$) of the quantization amounts in this method
\subsection{Iterative Quantization with $l_1$ Regularization}
One major drawback of algorithm \ref{alg:l1Quantization} and the improved $l_1 + l_2$-based algorithm is that they could not explicitly indicate the number of demanded distinct values (quantization amounts). To obtain an algorithm capable to explicitly specify amount of distinct values, an iterative method is designed. The paradigm of the algorithm is straightforward: it starts with a small $\lambda_1$ value and gradually increase the quantities of it, until the amount of non-zero values of the optimized $\boldsymbol{\alpha}$ could be reduced to equal to or less than the required amounts. Specifically, at each iteration, the algorithm will firstly follow the procedure of algorithm \ref{alg:l1Quantization} with current $\lambda_1$. After obtaining the optimized $\boldsymbol{\alpha}^{*}_{t}$ of the $t$-th iteration, it will be put back to the target function and the $\boldsymbol{\alpha}^{*}_{t+1}$ will be obtained with the algorithm \ref{alg:l1Quantization} at the $(t+1)$-th iteration.\par
The iterative quantization method could be described as algorithm \ref{alg:l1QuantizationIterative}.
\begin{algorithm}
\caption{Quantization with Iterative $l_1$ Optimization}\label{alg:l1QuantizationIterative}
\hspace*{\algorithmicindent} \textbf{Input: }\text{Original vector $\boldsymbol{w}$, Desired number of distinct value $l$} \\
\hspace*{\algorithmicindent} \textbf{Output: }\text{Quantized vector $\boldsymbol{w}^{*}$} 
\begin{algorithmic}[1]
\State $\boldsymbol{\hat{w}} \gets \texttt{unique}(\boldsymbol{w})$
\State $\lambda_1^{0}$ with a small number
\State $\Delta\lambda \gets \lambda_1^{0}$
\While{$||\boldsymbol{\alpha}||_{0}>l$}
\State $\lambda_1^{t} = \lambda_1^{0} + (t-1)\Delta\lambda$
\State Optimize target function \ref{l1normReguliriseOpt} with Coordinate Descent, with $\lambda_1 \gets \lambda_1^{t}$, $\boldsymbol{\alpha}^{0}_{t} \gets \boldsymbol{\alpha}^{*}_{t-1}$
\State Retrieve $\boldsymbol{V}^{*}_{t}$ with equation \ref{NpStarRetrievalMethod}
\State Compute $\boldsymbol{\hat{\alpha}}^{*}_{t}$ with equation \ref{PostLassoLeastSquareSolution}
\State Compute $\boldsymbol{\alpha}^{*}_{t}$ vector with equation \ref{FinalAlphaComputation}
\EndWhile
\State Compute the desired $\boldsymbol{w}^{*}$ vector with equation \ref{QuantizedVectorComputing} and final $\boldsymbol{\alpha}^{*}_{T}$
\end{algorithmic}
\end{algorithm}
\subsection{Clustering-based Least Square Sparse Optimization}
Algorithm \ref{alg:l1QuantizationIterative} could provide quantization results with given amount of distinct values $l$. However, since the algorithm could be sensitive to the change of $\lambda_1$, in practice it might fail to optimize to exact $l$ values but provide $\hat{l}<l$ values instead. Similarly, for the algorithm with equation \ref{l0normReguliriseOpt}, we could only set the upper bound of the amount of distinct values and there are no guarantees for how many distinct values will finally be produced. To further improve the capacity of quantization algorithm, here we discuss a general target that could produce definite amount of values with least square form, and design a basic method based on the combination of k-means clustering and least square optimization.\par
Suppose we want to construct a vector with $l$ distinct values now, and here we directly set the parameter vector $\boldsymbol{\alpha}$ to a vector with $l$ entries ($[l \times 1]$ shape). And now we need to use a transformation matrix to transform the $l$-value vector into a $[m \times 1]$ vector while maintaining the values constructed. One possible scheme could be to use a $m \times p$ transformation matrix $\boldsymbol{E}$ with one-hot encoded at each row. Under this scheme, the optimization target will be as following:
\begin{equation}\label{directOptmizationTarget}
\begin{aligned}
& \underset{\boldsymbol{E},\boldsymbol{\alpha}}{\text{min}}
& & ||\boldsymbol{\hat{w}} -\boldsymbol{E}\boldsymbol{V^{*}}\boldsymbol{\alpha}||_{2}^{2} \\
& \text{subject to}
& & ||\boldsymbol{E}_{i}||_0 = 1, \forall i = 1,2,...,m\\
&&& ||\boldsymbol{E}_{i}||_1= 1, \forall i = 1,2,...,m\\
\end{aligned}
\end{equation}
The constraint of $\boldsymbol{E}$ in equation \ref{directOptmizationTarget} means that for each row in the matrix, it will be 1 if we want the corresponding value to be belonging to this cluster; otherwise, it will 0. An alternative expression of the matrix will be:\\
\begin{equation}
\label{EMatforCluster}
\boldsymbol{E}_{ij} = 
\begin{cases}
1, & I(\boldsymbol{\hat{w}_{*}}) = j \\
0, & \text{else}
\end{cases}
\end{equation}
here $I(\hat{W}_i)$ means the group(cluster) which the $i$th value belongs to. The optimization will be difficult to perform with two optimization variables and a discrete geometry. Here, we propose a simple approximation to deal with this problem: one can first perform clustering (e.g. k-means) to obtain $I(\boldsymbol{\hat{w}}) = K(\boldsymbol{\hat{w}})$ to approximate matrix $\boldsymbol{E}$. Then the target function can be further transferred into the following expression:\\
\begin{equation}\label{directOptmizationTargetCluster}
\begin{aligned}
&\min_{\boldsymbol{\alpha}} ||\boldsymbol{\hat{w}}-\boldsymbol{E}\boldsymbol{V^{*}}\boldsymbol{\alpha}||_{2}^{2}\\
=&\min_{\boldsymbol{\alpha}} ||\boldsymbol{\hat{w}}-\boldsymbol{\hat{V}^{*}}\boldsymbol{\alpha}||_{2}^{2}
\end{aligned}
\end{equation}
And notice that since the 'index of non-zeros' of the algorithm is obtained through clustering, the rank of the $\boldsymbol{\hat{V}^{*}}$ is no longer a problem of concern. Hence, one could simply compute the value $v=\text{mean}(\boldsymbol{\hat{w}})$ to fill all of the non-zero entries. Based on the above settings and equation \ref{EMatforCluster} and \ref{directOptmizationTargetCluster}, the matrix $\boldsymbol{\hat{V}^{*}}$ would be the follows:\\
$$
\boldsymbol{\hat{V}^{*}} =
\begin{bmatrix}
v & 0 & 0 & \cdots & 0\\
v & 0 & 0 & \cdots & 0\\
\cdots & \cdots & \cdots & \cdots & \cdots \\
v & 0 & 0 & \cdots & 0\\
v & v & 0 & \cdots & 0\\
\cdots & \cdots & \cdots & \cdots & \cdots \\
v & v & v & \cdots & v
\end{bmatrix}
$$
The optimization of equation \ref{directOptmizationTargetCluster} is a typical linear regression problem and could be solved in closed form with polynomial $O(l^{2}m+l^{3})$ time complexity or even faster with approximations \cite{dhillon2013LeastSquareApproximation}. By taking derivatives and set it to 0, we could obtain the solution:\\
\begin{equation}\label{clusteringExactQuantizationSolution}
\boldsymbol{\alpha} = (\boldsymbol{\hat{V}^{*}}^{T}\boldsymbol{\hat{V}^{*}})^{-1}\boldsymbol{\hat{V}^{*}}^{T}\boldsymbol{\hat{w}_{*}}
\end{equation}
The algorithm of the clustering-based least square method could be given as algorithm \ref{alg:EaxctQuantizationLS}. One interesting point is, from the perspective of clustering methods, algorithm \ref{alg:EaxctQuantizationLS} could be viewed as an improvement of k-means clustering quantization. In conventional clustering-based quantization algorithm, the representation of a certain cluster of values is simply given as the mean of the cluster. In contrast, for the proposed algorithm, it alternatively computes the value of the cluster that produce the smallest least square distance from the original.\par
Notice that there should exist multiple schemes to solve the optimization problem proposed by equation \ref{directOptmizationTarget}, and the method proposed here is only a basic solution. The exploration of solving this task could be one of our future research focuses. 
\begin{algorithm}
\caption{Quantization with K-means-based Least Square}\label{alg:EaxctQuantizationLS}
\hspace*{\algorithmicindent} \textbf{Input: }\text{Original vector $\boldsymbol{w}$, Desired number of distinct value $l$} \\
\hspace*{\algorithmicindent} \textbf{Output: }\text{Quantized vector $\boldsymbol{w}^{*}$} 
\begin{algorithmic}[1]
\State $\boldsymbol{\hat{w}} \gets \texttt{unique}(\boldsymbol{w})$
\State Perform k-means with $l$ clusters, get model $\boldsymbol{K^{*}(\cdot)}$
\State Apply $\boldsymbol{K^{*}(\boldsymbol{\hat{w}})}$ to get the prediction of each data
\State Fill the corresponding columns with 1 for matrix $\boldsymbol{E}$ according to equation \ref{EMatforCluster}
\State Optimize $\boldsymbol{\alpha}$ according to equation \ref{directOptmizationTargetCluster}. The base value of $v$ could be $v = \text{mean}(\boldsymbol{\hat{w}})$
\State Compute the desired $\boldsymbol{w}^{*}$ vector with equation $\boldsymbol{w}^{*} = \boldsymbol{\hat{V}^{*}}\boldsymbol{\alpha}^{*}$
\end{algorithmic}
\end{algorithm}

\subsection{Time Complexity Analysis}
As it has been mentioned above, the proposed algorithm cannot outperform k-means-based quantization method consistently in every case. In fact, as we will see in this section, if the number of iterations to compute k-means and sparse least square are asymptotically the same, the proposed quantization method will run with a worse time complexity. However, in practice, k-means often takes significantly more iterations to converge, and to avoid out-of-the-range values and empty sets, multiple trails are usually required. In this sense, the proposed least square-based algorithm will have a more favorable time complexity. \par 
With (block) Coordinate Descent, the time complexity of convex $l_{1}$ Lasso regression is $O(t_{l}mn)$, where $m$ and $n$ are the magnitude and number of dimensions of data, respectively, and $t_{l}$ is the number of iterations. Under the setting of algorithm \ref{alg:l1Quantization}, since the matrix to be optimized is $m \times m$ (where $m$ is the length of $\boldsymbol{\hat{w}}$), the complexity will be $O(t_{l}m^{2})$. As we have seen in section \ref{subsubsec:convergence}, the target function will converge globally and linearly under Coordinate Descent optimization, and the number of iterations to expect is usually less than those of k-means. In our experiments, as one can observe from section \ref{sec:experiment}, the optimum can usually be found in acceptable number of iterations.\par
With the popular Lloyd's method, the time complexity of k-mean algorithm is $O(t_{k}kTm)$, where $t_{k}$ is the number of iterations; $k$ is the number of cluster centroids (and the desired number of distinct values in scalar quantization); $m$ is the length of $\boldsymbol{\hat{w}}$; and $T$ is the times of trials to guarantee convergence. By comparing the two complexities, one can observe that if there is $t_{k}kT \in \Omega(t_{l}m)$, the k-means-based method will be asymptotically running in a higher complexity, and the complexity of the proposed method will become more preferable. This case will unlikely to happen when $k$ is small, but will become common when we have $k\in \theta(m)$. That is to say, if moderate to large number of post-quantization values(high-resolution results) is desired in the quantization task, the proposed algorithm will converge faster than k-means. This type of quantization requirements are not unusual in engineering applications: for instance, sometimes it would be desirable to reduce the number of distinct values to the nearest $2^k$ to reduce memory cost yet preserve most of the information. Consequently, in such cases, the proposed algorithm will be much more favorable than existed k-means clustering-based one.
\section{Experimental Results}
\label{sec:experiment}
To verify the rationality and effectiveness of the proposed methods, three types of data, namely neural network fully-connected layer weight matrix, MNIST image, and artificially generated data sampled from different distributions, are employed to obtain experimental results for illustrations and analysis. The performances are evaluated mostly based on quantization information loss and time-consumption. The information loss is denoted by $l_2$ loss between the original vector and the quantized vector for MNIST image and artificially-generated data, and by post-quantization recognition accuracy for the neural network compression. Notice that in some certain scenarios, a high $l_2$ loss may not necessarily mean a deficient performance. For example, in image quantization, the $l_2$ loss could be dominated by few values far away from the original, and the image with higher $l_2$ loss might actually possess an overall more favorable quality. Thus, for the quantization of MNIST images, the post-quantization results are plotted as images in figure \ref{fig:MNISTcomparison} to assistant one to evaluate the performances from a human intuition. \par
Another point to notice is that in the experiments of MNIST and artificially-generated data, the post-quantization outputs are processed by a 'hard-Sigmoid' function before they are utilized to compute the $l_2$ information loss. The 'hard-Sigmoid' function is denoted as follows:
\begin{equation}
H(x,a,b) =
\begin{cases}
a, & x\leq a\\
x, & a<x<b\\
b, & x\geq b
\end{cases}
\end{equation}
Where $a$ and $b$ are the 'floor' and 'ceiling' of the range of values. The reason for this function to be implemented is that in many situations, the quantization results must lie in a certain range. For example, MNIST quantization values must be in $[0,1]$, otherwise it will not be recognized in practical image storage/displaying systems. Applying the function could avoid out-of-range values that might reduce the $l_2$ loss in a prohibited way. \par
Major experiments in this section involve the comparison between selected existing methods, including k-means clustering, Mixture of Gaussian quantization, and quantization based on a recently proposed data transformation clustering\cite{azimi2017novelClustering}, and the proposed algorithms, including the $l_1$ quantization method (equation \ref{l1normReguliriseOpt}), $l_1$ with least square method (algorithm \ref{alg:l1Quantization}), and clustering-based least square method (algorithm \ref{alg:EaxctQuantizationLS}). For the experiments of neural network weights and artificially-generated data with specific distributions, all the algorithms listed except the $l_0$ optimization-based one are tested (since it could not consistently procude stable results). In contrast, in the experiment of MNIST images, we focused on comparing the proposed methods, including the $l_0$-based algorithm, with k-means clustering. In addition, the performance comparison between sole $l_1$-based and $(l_1 + l_2)$-based quantization is examined with a separate optimization program implemented by the authors. Furthermore, the performances of $l_0$-based optimization method (equation \ref{l0normReguliriseOpt}) is implemented and tested with the previously-mentioned $l_0$ optimization software\cite{l0Learn2018Hazimeh} based on R. The $l_1$ optimization and k-means-based methods are accomplished via Lasso and k-means in Sk-learn\cite{scikit-learn}, respectively, and the codes are both optimized to the possible extent as far as the authors concern. One might concern that the multiple times of running optimization(10 times by default) in the k-means of Sk-learn could lead to an unfair comparison for the time consumption. However, running multiple times is the standard practice for K-means to produce reliable results. Furthermore, for fuzzy c-means clustering, \cite{wen2011hard} illustrates that it will take longer time than k-means (hard c-means), yet the performance are not significantly better. Thus, fuzzy c-means quantization is not included in the experiments. \par
Experimental results illustrate that in general, 1. The $l_1$-based quantization method will lead to a higher $l_2$ information loss comparing to K-means clustering, but the running time can be considerably reduced for medium-size data. Meanwhile, applying quantization with Mixture of Gaussian has slightly worse performance for neural network weights comparing with K-means but approximately same results for the artificially-generated data, while employing the novel data transformation-based clustering will produce a similar performance with K-means for the neural-network weights but worse performance for the artificially-generated ones; 2. After applying least square to $l_1$-based sparse quantization method, the performance can be much more competitive and the information loss will be in the same level of k-means, while the running time is still significantly quicker than K-means; 3. The clustering-based least square method can perform slightly better than K-means, and it does not take significant longer running time; 4. the combined $l_1$+$l_2$ optimization can induce fewer distinct values (quantization amounts) with the same $\lambda_1$ of sole $l_1$ method, but the algorithm is sensitive with the value of $\lambda_2$; and finally, 5. $l_0$-based quantization method (under the optimization algorithm provided in \cite{l0Learn2018Hazimeh}) can provide good performance within acceptable running time, but it could not universally produce quantization results (some amounts of quantization amounts are irretrievable) and the optimization could fail under some circumstances (especially when the demanded quantization amount is large).
\subsection{Neural Network Weight Matrix}
\begin{figure}
\centering
\includegraphics[width=0.9\textwidth]{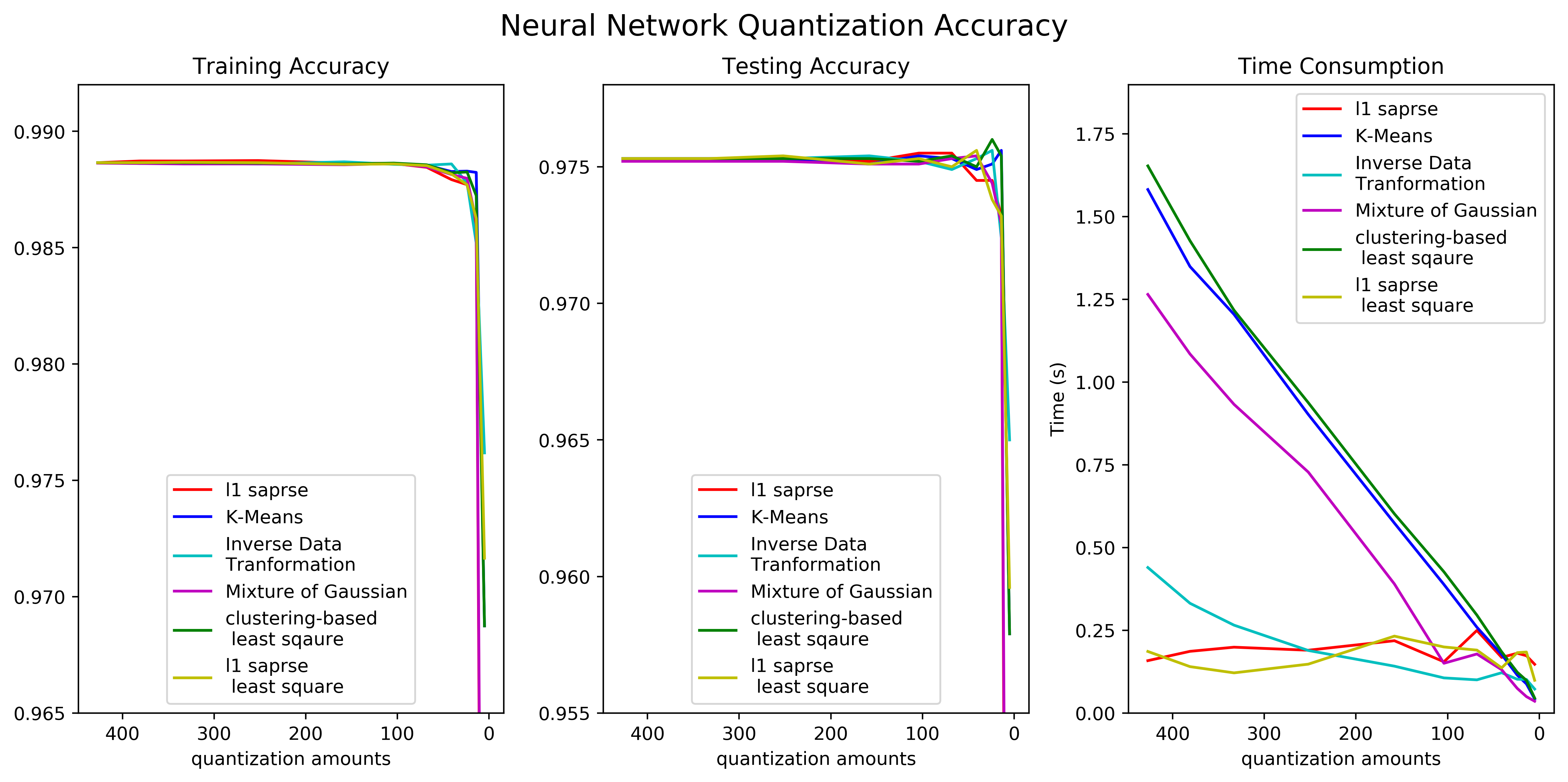}
\caption{Post-quantization accuracy on training and testing data with respect to quantization amounts, and their running time. The x-axis for the plots stands for quantization amounts(number of quantization), and the y-axis stands for accuracy for the first two plots and time in seconds for the third plot.}
\label{fig:NNAccuracyHolistic}
\end{figure}

\begin{figure}
\centering
\includegraphics[width=0.8\textwidth]{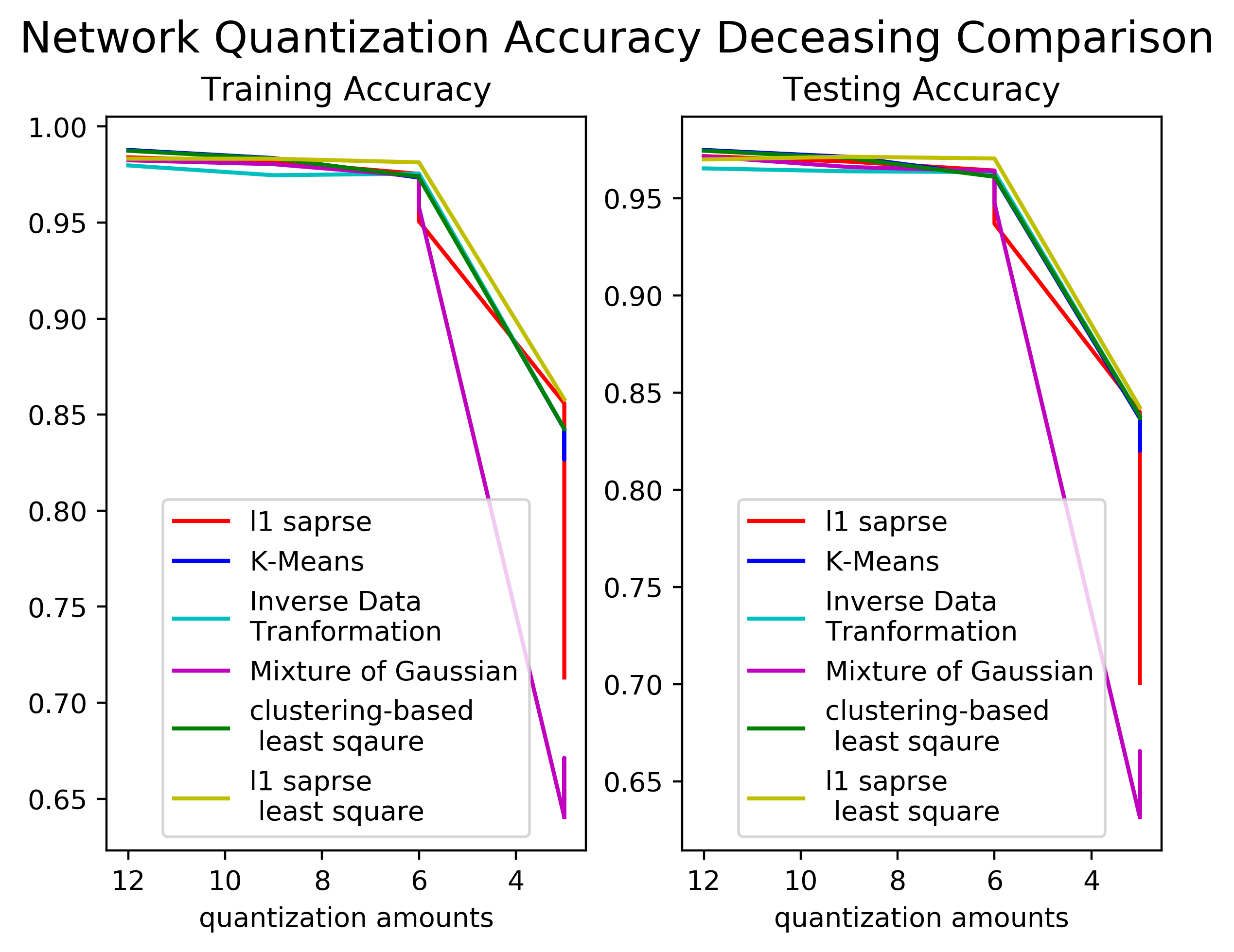}
\caption{Post-quantization accuracy on training and testing data with respect to quantization amounts, focusing on the area where accuracy drops significantly. The x-axis stands for number of clusters, the y-axis stands for accuracy.}
\label{fig:NNAccuracyZoomingIn}
\end{figure}

To test the effectiveness of our methods on neural network quantization (which is also the original problem inspired this paper), a 5-layer  $784-256-128-64-10$ fully-connected network for MNIST image recognition is introduced. The network is trained with stochastic gradient descent and the original accuracy on training and testing data are $98.86\%$ and $97.53\%$, respectively. In the experiments, the last layer ($64-10$) matrix is processed by the quantization method and the weights are replaced by the post-quantization matrix. Figure \ref{fig:NNAccuracyHolistic} illustrates the change of accuracy on training and testing data with respect to different number of quantization values (cluster numbers) for $l_1$, $l_1$ least square, k-means and cluster-based least square methods. In addition, the running time of each method is demonstrated in the figure. And since the accuracy of MNIST recognition is fairly robust against quantization, figure \ref{fig:NNAccuracyZoomingIn} further provides a figure zooming in the area that the accuracy starts to drop with a higher precision of quantization amounts. 
\par
From the figures it can be found out that the proposed sparse regression-based methods can provide competitive performances for the last-layer quantization of the neural network. The pure $l_1$ sparse regression provides a slightly more deficient performance than other proposed methods, but in most cases the deficiency is negligible and it can still outperform Mixture of Gaussian-based quantization. Comparing to the time-consuming k-means based methods, the proposed methods can provide an alternative solution with much quicker running time. In addition, if the least square method is applied to optimize the $\boldsymbol{\alpha}$ of $l_1$ as algorithm \ref{alg:l1Quantization} does, the algorithm will be able to provide no less competitive results than k-means method does in terms of accuracy, while the running time will remain at a low level. The clustering-based least square exact method can provide the overall optimal performance, especially in the area approaching to accuracy decrements. And the additional time consumed by solving least square on the top of k-means is negligible. \par
\begin{figure}
\centering
\includegraphics[width=1.0\textwidth]{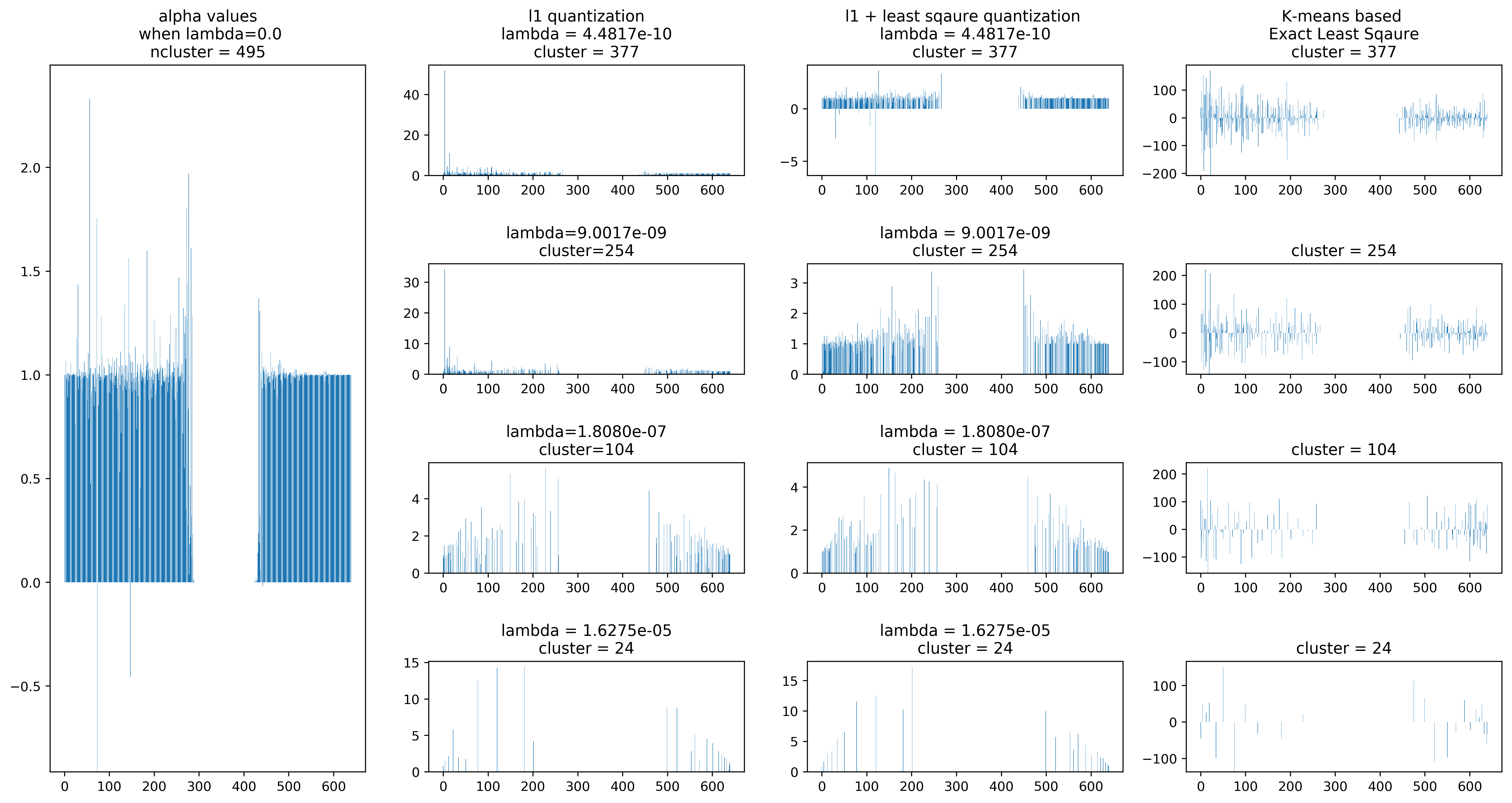}
\caption{The distribution of $\boldsymbol{\alpha}$ weights for neural network last-layer quantization}
\label{fig:NNQuantizationAlphaComparison}
\end{figure}

Figure \ref{fig:NNQuantizationAlphaComparison} shows the $\boldsymbol{\alpha}$ values of the neural network last-layer quantization with different level of sparsity (quantization amounts). The full-column plot on the left is the weights solved by least square without sparsity. It could be found that even for the least square solution without any additional regularization term, some of the values of $\boldsymbol{\alpha}$ still hit the values around or equal to 0. The plots of the rest three columns represent the $\boldsymbol{\alpha}$ values of $l_1$ without least square, $l_1$ with least square and k-means based exact value methods, respectively. The trick behind the third plot is that the values of the dense vector are assigned to the starting index of each 'batch' of same-values, which can form a equivalent illustration to a sparse $\boldsymbol{\alpha}$ parameter. \par
From the plots of $l_1$-based quantization, it could be observed that almost all the $\alpha_{i}$ are positive. This result is consistent with the characteristics of Coordinate Descent, which utilize a shrinkage operator and decreases small positive values to $0$. The major difference between sparse least square methods and clustering-based method is that the latter tends to produce positive and negative values in roughly equivalent amounts. However, despite the differences between values of vectors, the two types of methods share a 'central zero area' for index around 300-400. This implies that the proposed least square-based methods can capture the geometric information of the vector similar to clustering-based methods, while the time complexity is significantly lower. Furthermore, a most-positive $\boldsymbol{\alpha}$ vector is more consistent with the configuration of the $\boldsymbol{V}$ matrix. \par
\begin{figure}
\centering
\includegraphics[width=0.8\textwidth]{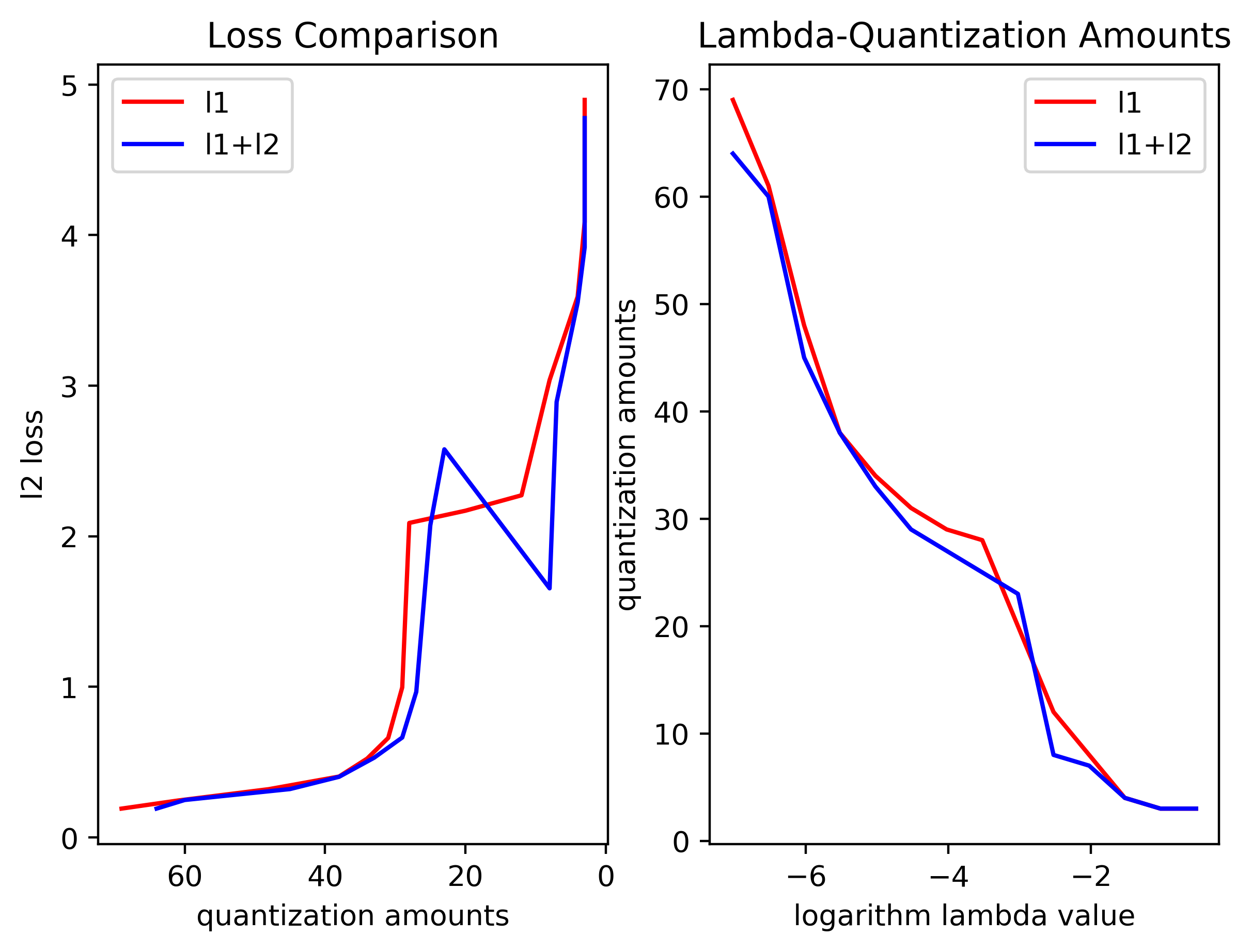}
\caption{The accuracy with respect to $\lambda_1$ values for sole $l_1$ and $l_1+l_2$ optimization, respectively. The value of $\lambda_2$ is set to $|\lambda_2| = 4*10^{-3}*\lambda_1$.}
\label{fig:NNQuantizationl1l2Comparison}
\end{figure}

Moreover, to illustrate the effects of replacing $l_1$ with $l_1 + \text{(negative) }l_2$ optimization, the performance of $l_1$ and $l_1 + l_2$ optimization on the last-layer neural network data is illustrated in figure \ref{fig:NNQuantizationl1l2Comparison} with a coordinate descent optimization method implemented separately. Neither of the $\boldsymbol{\alpha}$ weights in the illustrated plots is optimized with least square. From the figure it could be observed that the $l_1+l_2$ method could in general lead to fewer quantization amounts for the same $\lambda_1$ value, while produce a smaller $l_2$ loss comparing to the original. The experimental result also verifies the argument in section 3.3. However, despite the favorable performance, the algorithm could be sensitive to the value of $\lambda_2$, and it could be numerically very unstable if the value of $\lambda_2$ is too large or too small. To improve the tolerance of $\lambda_2$ in this algorithm could be a point of exploration in the future. \par
And finally, as for the $l_0$ quantization method, it could not find any non-trivial solution under the optimization method of \cite{l0Learn2018Hazimeh}, which indicated the drawbacks of numerical unreliability of $l_0$-based method.
\subsection{MNIST Image Quantization}
\begin{figure}
\centering 
\includegraphics[width=0.9\textwidth]{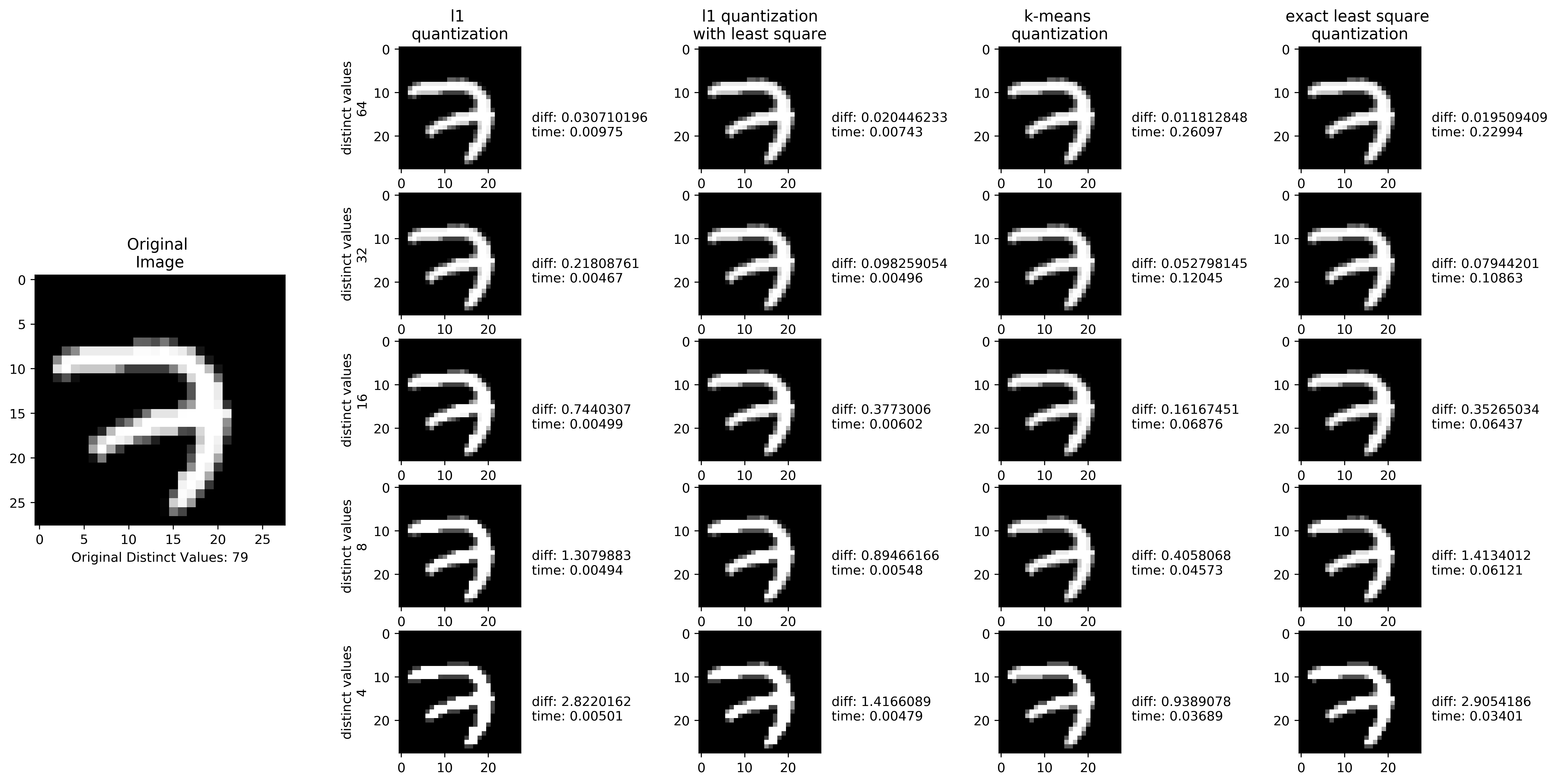}
\caption{MNIST quantization comparison for $l_1$ method, $l_2$ and least square method, k-means method, k-means-based least square method}
\label{fig:MNISTcomparison}
\end{figure}

Quantization could be used in image processing to reduce the number of values and space complexity. In this paper, a MNIST-digit image is chosen as an example to show the performance of image quantization of the proposed methods. The performances of two types of $l_1$-based algorithms and two kinds of clustering-based methods are illustrated and compared in Fig$.$\ref{fig:MNISTcomparison}. From the figure we could find out that K-means and the clustering-based least square optimization can provide the best performances in general, and there is no significant differences in terms of execution time. $l_1$ with least square optimization of $\boldsymbol{\alpha}$ values can provide less norm difference loss than using $l_1$ solely. Meanwhile, in terms of running time, the $l_1$-based optimization approaches can provide significant advantages over K-means-based methods. Another remark of the MNIST quantization is that the K-means methods sometime provide out-of-range values (not in the interval $[0,1]$) when the number of clusters are large, and the reason can be attributed to bad initialization. However, for the least-square optimization methods, this problem does not happen at least in the MNIST circumstance.\par
\begin{figure}[t]
\centering
\includegraphics[width=0.8\textwidth]{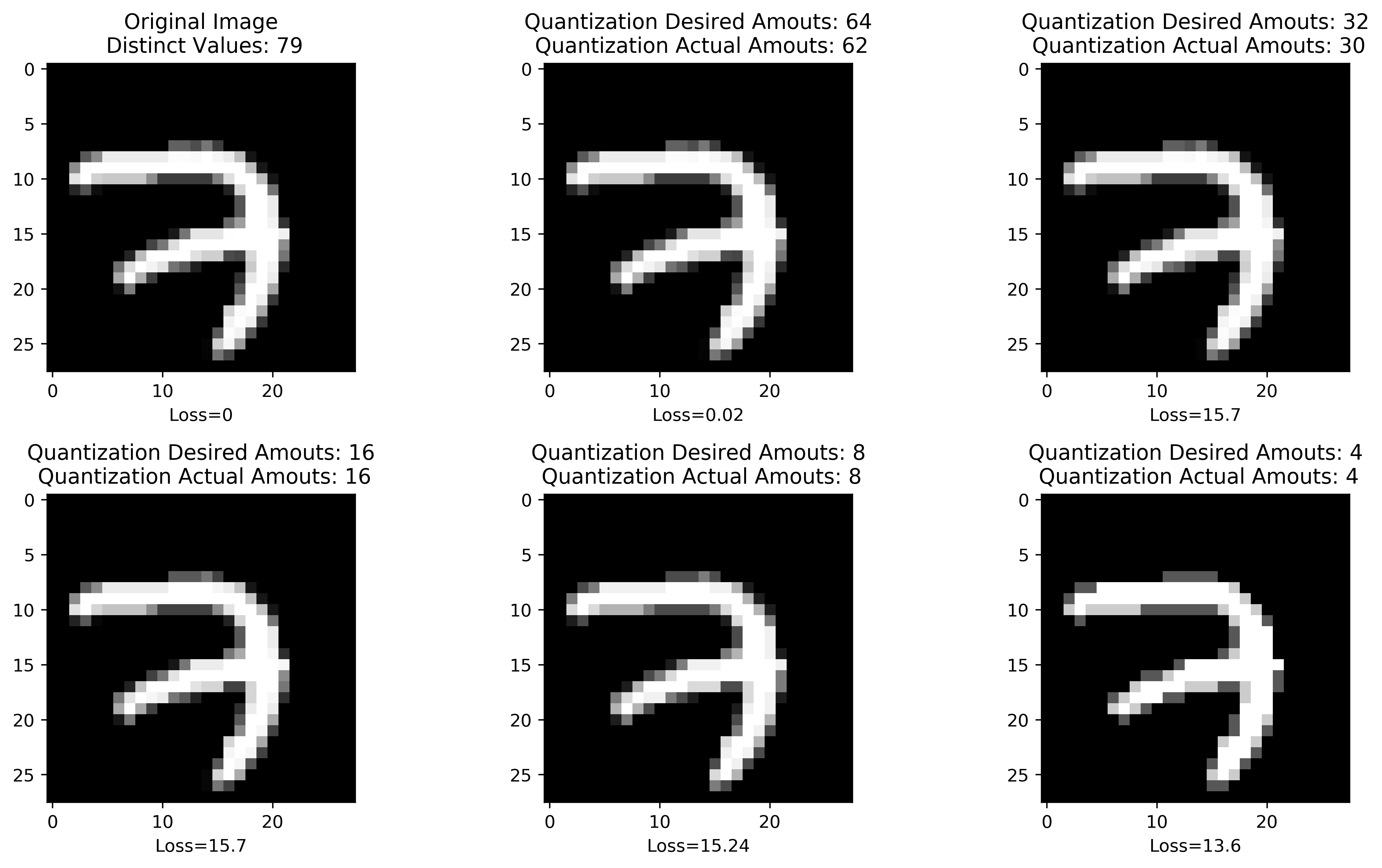}
\caption{MNIST quantization results on the $l_0$-based method}
\label{fig:MNISTL0Result}
\end{figure}
The quantization result of $l_0$ method is shown separately in figure \ref{fig:MNISTL0Result}. It could be observed from the figure that the qualities of images are in general high and the $l_2$ loss is competitive. However, the problem of 'not universal' is also very significant from the figures: in many cases, the algorithm could only find the largest possible quantization amounts smaller than the given value. In addition, the algorithm often fail to find solution when the required quantization amount is large.
\subsection{Artificially-Generated Data with Specific Distributions}
\begin{figure}
\centering 
\includegraphics[width=0.8\textwidth]{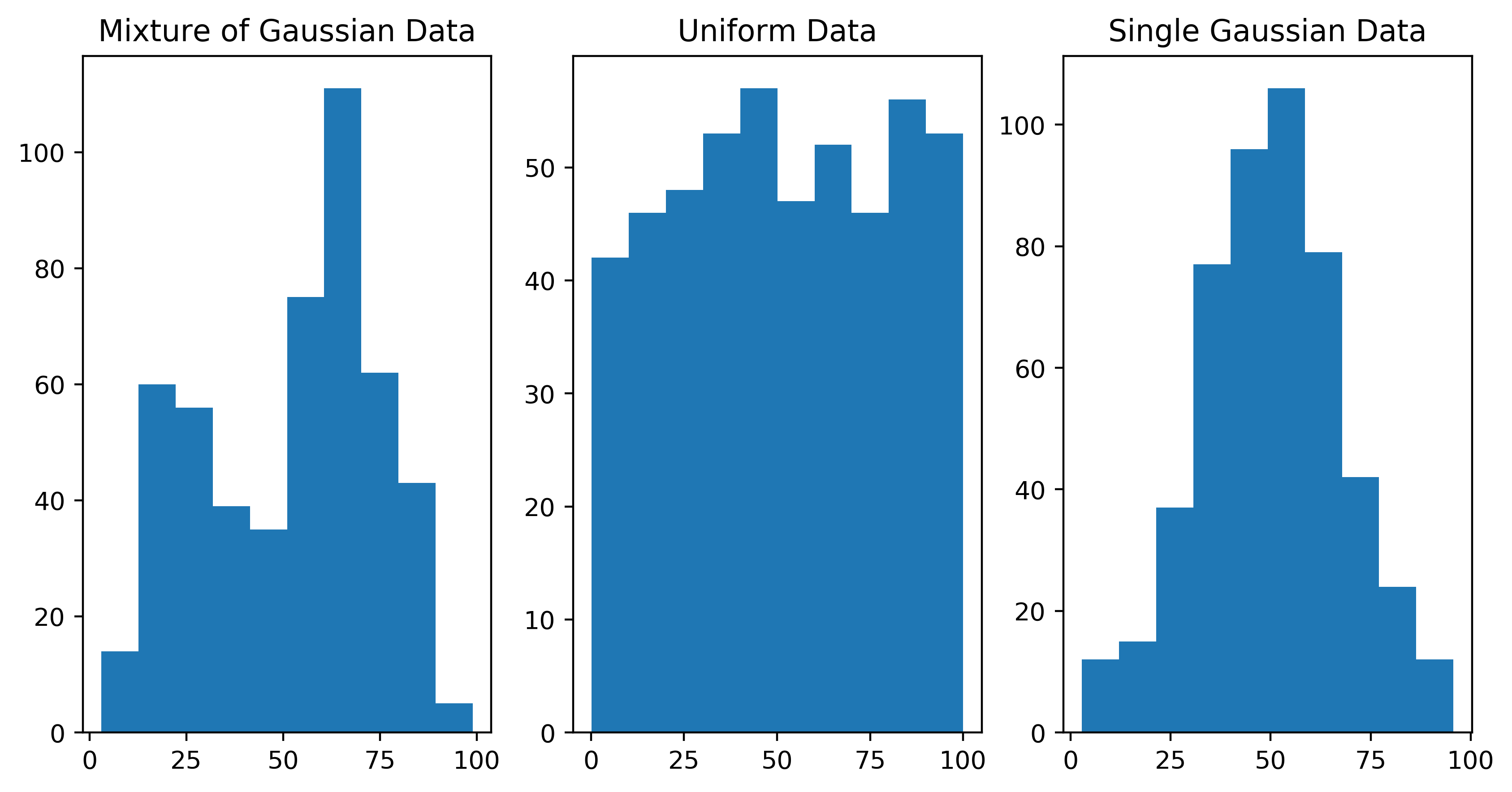}
\caption{The distribution of the 3 types of artificially-generated data}
\label{fig:ArtificialDataIllustration}
\end{figure}

\begin{figure}
\centering 
\includegraphics[width=1.0\textwidth]{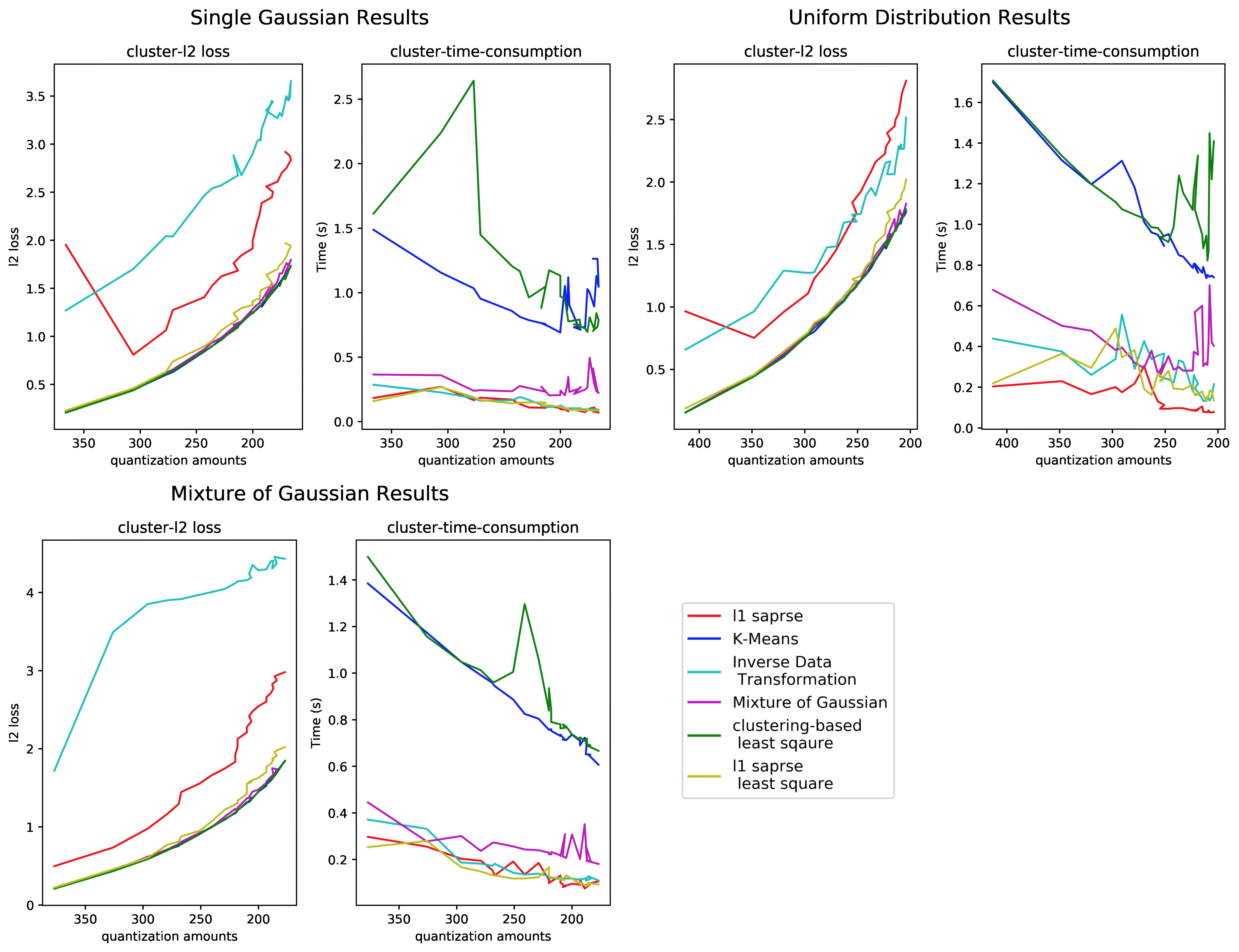}
\caption{Quantization results of artificially-generated data. For each subplot, the left is the norm loss figure, and the right one is the running time. The x-axis stands for clusters, and the y-axis stands for l2 loss for the left figures, and for time in seconds for the right ones.}
\label{fig:PerformanceartificalGeneratedData}
\end{figure}

To test the performance of the proposed algorithms on data of different distributions, three types of distributions, namely Mixture of Gaussian, Uniform and Single Gaussian, are employed to generate 500 samples from each of them. The samples are constrained in the range of $[0,100]$, and the distributions of the data we used could be shown as figure \ref{fig:ArtificialDataIllustration}. In practice, these three types of distributions could describe most cases of 1-d data characteristics.\\
The experimental results for different quantization methods based on the above 3 types of data is shown in figure \ref{fig:PerformanceartificalGeneratedData}. From the figure, it could be observed that the information loss deficiency of $l_1$ without least square is more significant comparing to those for neural network and MNIST image. However, considering the running time reduced by the algorithm, the overall performance could still be regarded as merited. Also, if least square is employed to optimize the values of the vector, the information loss of $l_1$ approach will be only slightly higher than k-means based methods optimization, while the run-time will be still of great advantage comparing to the k-means branch of methods.\\
And finally, for the $l_0$ quantization method, again in the experiments it could not provide meaningful results for the quantization of the artificially-generated data. This problem further demonstrates the issue of using $l_0$ optimization despite its favourable information loss: getting the exact solution of $l_0$ is NP-hard, and with approximation optimization, there is a risk of failure in getting the results.
\section{Conclusion}
This paper proposed several sparse least square-based algorithms to better accomplish the task of scalar quantization. The characteristics and computational properties of the proposed algorithms are examined, and the advantages, drawbacks, and the advantageous scenarios of each of them are analyzed. The algorithms are implemented and tested under the scenarios of neural network weight, MNIST image and artificially-generated data respectively, and the results are demonstrated and analyzed. Experimental results shows that the proposed algorithms have competitive performances in terms of information loss/preservation, and the favorable properties in running time could make the $l_1$-based algorithms especially useful when processing large batch of medium-size data and the number of post-quantization values are not far from the original. \par
The paper made the following major contributions: Firstly, it proposed several novel quantization methods with competitive information preservation ability and much more favorable time complexity; Secondly, the paper innovated the pioneering work of using least square optimization to solve the quantization problem, which could bring huge research potentials in the area; And thirdly, the algorithms proposed in the paper provide broader options for engineering practice, especially when the purpose of performing quantization is to restrict the number of distinct values to a high-resolution level, and when the number of post-quantization values is still large. \par
In the future, the authors intend to continue to explore quantization algorithms based on the idea initiated in the paper. One major problem of interests will be to extend the quantization method into high-dimensional vector scenarios by adopting novel target function and optimization method.
\section{Acknowledgement}
The authors would like to thank Mr. Feng Chen from Northwestern Polytechnical University, China, and Mr. Shixiong Wang from National University of Singapore, Singapore. Mr. Chen and Mr. Wang provided the authors many useful comments for the algorithms and programs of the research.
\section{Declaration of interest}
The authors declare no conflicts of interest.
\clearpage

\bibliographystyle{model1-num-names}
\bibliography{sample}

\end{document}